\pdfoutput=1

\documentclass[11pt]{article}

\usepackage[]{ACL2023}

\usepackage{amsmath}
\usepackage{times}
\usepackage{latexsym}

\usepackage[T1]{fontenc}

\usepackage[utf8]{inputenc}

\usepackage{microtype}

\usepackage{inconsolata}

\usepackage{multicol}
\usepackage{multirow}
\usepackage{xcolor}
\usepackage{calc}
\usepackage{amsfonts}
\usepackage{graphicx}
\usepackage{tabularx}

\usepackage{booktabs}
\title{LambdaKG: A Library for Pre-trained Language Model-Based\\ Knowledge Graph Embeddings}

\author{
Xin Xie\textsuperscript{\rm 1}\thanks{\quad Equal contribution and shared co-first authorship.}, 
Zhoubo Li\textsuperscript{\rm 1}\footnotemark[1], 
Xiaohan Wang\textsuperscript{\rm 1}\footnotemark[1], 
Zekun Xi\textsuperscript{\rm 1}, 
\textbf{Ningyu Zhang}\textsuperscript{\rm 1 \thanks{\quad Corresponding author.}} \\
\textsuperscript{\rm 1}  Zhejiang University \\
\texttt{\{xx2020,zhoubo.li,wangxh07,zhangningyu\}@zju.edu.cn},\\
\url{https://zjunlp.github.io/project/promptkg/}
}

\def\ours{\textbf{LambdaKG}}
\begin{document}
\maketitle
\begin{abstract}

Knowledge Graphs (KGs) often have two characteristics: heterogeneous graph structure and text-rich entity/relation information. Text-based KG embeddings can represent entities by encoding descriptions with pre-trained language models, but no open-sourced library is specifically designed for KGs with PLMs at present. In this paper, we present {\ours}, a library for KGE that equips with many pre-trained language models (e.g., BERT, BART, T5, GPT-3) and supports various tasks (e.g., knowledge graph completion, question answering, recommendation, and knowledge probing). {\ours} is publicly open-sourced\footnote{Code: \url{https://github.com/zjunlp/PromptKG/tree/main/lambdaKG}}, with a demo video\footnote{Video: \url{http://deepke.zjukg.cn/lambdakg.mp4}} and long-term maintenance.

\end{abstract}

\section{Introduction}

Knowledge Graphs (KGs) encode real-world facts as structured data and have drawn significant attention from academia, and industry \cite{zhang2022deepke}. 
Knowledge Graph Embedding (KGE) aims to project the relations and entities into a continuous vector space, which can enhance knowledge reasoning abilities and feasibly be applied to downstream tasks: question answering \cite{DBLP:conf/acl/SaxenaKG22}, recommendation \cite{DBLP:conf/kdd/ZhangJD0YCTHWHC21} and so on \cite{DBLP:conf/www/ChenZXDYTHSC22}.
Previous \textbf{embedding-based} KGE methods, such as TransE \cite{DBLP:conf/nips/BordesUGWY13}, involved embedding relational knowledge into a vector space and subsequently optimizing the target object by applying a pre-defined scoring function to those vectors.
A few remarkable embedding-based KGE toolkits have been developed, such as OpenKE \cite{DBLP:conf/emnlp/HanCLLLSL18}, LibKGE \cite{DBLP:conf/emnlp/BroscheitRKBG20}, PyKEEN \cite{DBLP:journals/jmlr/AliBHVSTL21}, CogKGE \cite{DBLP:conf/acl/JinMYHSWXCZ22} and NeuralKG \cite{neuralkg}.  
Nevertheless, these embedding-based KGE approaches are restricted in expressiveness regarding the shallow network architectures without using any side information (e.g., textual description).

\begin{figure*}[!t]

\centering 
\includegraphics[width=0.9\textwidth]{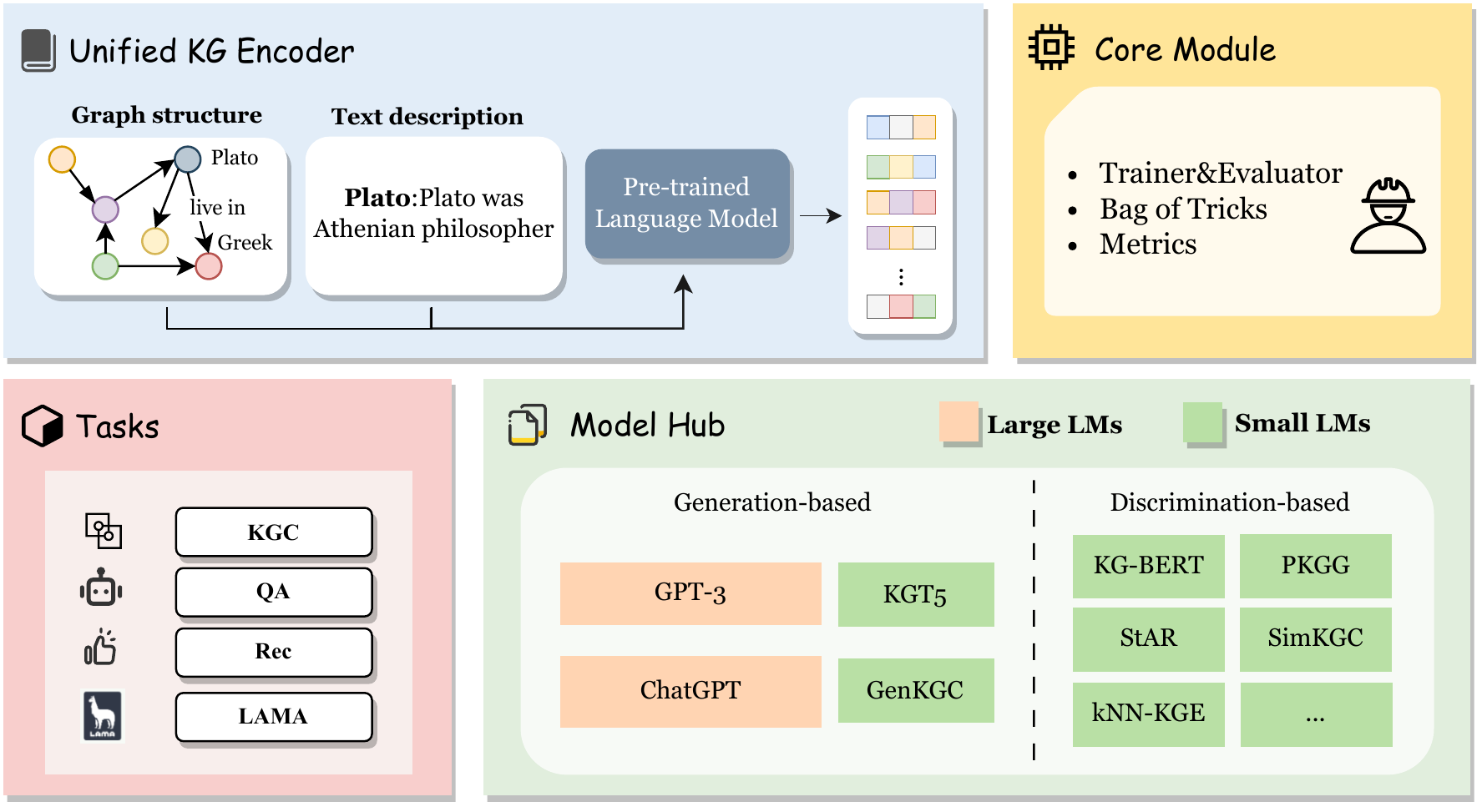} 
\caption{The architecture and features of \ours. 
}
\label{fig:framework}
\end{figure*}

By comparison with embedding-based KGE approaches, text-based methods incorporate available texts for KGE. 
With the development of Pre-trained Language Models (PLMs), many text-based models \cite{DBLP:conf/www/XieZLDCXCC22,DBLP:conf/acl/SaxenaKG22,kim-etal-2020-multi,markowitz-etal-2022-statik,DBLP:conf/coling/ChenWLL22,DBLP:conf/kdd/LiuZSCQZ0DT22} have been proposed, which can obtain promising performance and take advantage of allocating a fixed memory footprint for large-scale real-world KGs. 
Recently, large language models (LLMs) (e.g., GPT-3 \cite{DBLP:conf/nips/BrownMRSKDNSSAA20}, ChatGPT \cite{ChatGPT-OpenAI}) further demonstrated the ability to perform a variety of natural language processing (NLP) tasks without adaptation, providing potential opportunities of better knowledge representations.
However, there is \textbf{no comprehensive open-sourced library particularly designed for KGE with PLMs at present}, which makes it challenging to test new methods and make rigorous comparisons with previous approaches.

In this paper, we share with the community a pre-trained \textbf{LA}nguage \textbf{M}odel-\textbf{B}ase\textbf{D} libr\textbf{A}ry for \textbf{KG}Es and applications called {\ours} (MIT License), which supports various cutting-edge models.
Specifically, we equip {\ours} with both small PLMs, e.g., BERT \cite{DBLP:journals/corr/abs-1810-04805, DBLP:journals/corr/abs-1909-03193}, BART \cite{DBLP:conf/acl/LewisLGGMLSZ20, DBLP:conf/aaai/LiuW0PY21}, T5 \cite{DBLP:journals/jmlr/RaffelSRLNMZLL20, DBLP:conf/acl/SaxenaKG22}; and large PLMs, e.g., GPT-3 \cite{DBLP:conf/nips/BrownMRSKDNSSAA20}, ChatGPT \cite{ChatGPT-OpenAI}, by developing two major paradigms of discrimination-based and generation-based methods for KGEs.
\textbf{LambdaKG} supports factual and commonsense KGs with diverse tasks, including KG completion, question answering, recommendation, and knowledge probing (LAMA). 
We will provide maintenance to meet new tasks, new requests and fix bugs.

\section{System Architecture}

The overall features \& architecture of {\ours} are presented in Figure \ref{fig:framework}.
We will detail two major types of PLM-based KGE methods (discrimination-based and generation-based) with various PLMs.

Our design principles are:
1) \emph{Core Module with Unified KG Encoder}: \textbf{LambdaKG} utilizes a unified encoder to pack graph structure and text semantics, with convenient \texttt{Trainer\&Evaluator}, \texttt{Metric}, and \texttt{Bag of Tricks};
2) \emph{Model Hub}: \textbf{LambdaKG} is integrated with many cutting-edge PLM-based KGE models;
3) \emph{Flexible Downstream Tasks}: \textbf{LambdaKG} disentangles KG representation learning and downstream tasks.

\subsection{Core Module}

\subsubsection{Trainer\&Evaluator}

Typically, the training process with {\ours} can be decomposed into several distinct steps, such as the forward and backward passes (i.e., \texttt{training\_step}), logging of intermediate results (\texttt{log}), and model evaluation (\texttt{evaluate\_step}).
Our \texttt{Trainer} class provides a flexible and modular framework for training different types of models, with customizable functions to handle various tasks, such as computing the loss function and updating model parameters. 
Moreover, the \texttt{Trainer} class allows users to define their own plugins, which can be integrated seamlessly into the training pipeline to provide additional functionalities.

\subsubsection{Metric}
We design the \texttt{Metric} class to evaluate different models for various tasks.
Specifically, we use \texttt{hits@k} with k values of 1, 3, 10 and \textit{mean rank (MR)} as the evaluation metrics.
Hits@k measures the proportion of correct predictions among the top k-ranked results, while MR calculates the average rank of the correct answer.
We also implement \texttt{BLEU-1} score to evaluate the commonsense KG completion tasks following \citet{DBLP:conf/aaai/HwangBBDSBC21}.

\subsubsection{Bag of Tricks}
All models in the {\ours} are based on PLMs, and we equip a bag of tricks of training techniques to improve their performance.
In particular, we employ different pluggable modules such as \texttt{label smoothing} and \texttt{exponential moving average} to assist in the training of models.
We implement \texttt{early stopping} and \texttt{fast run} modules to prevent overfitting with small data by introducing early stopping and automatic verification mechanisms. 
Furthermore, we integrate an off-the-shelf \texttt{Top-k negative sampling strategy} to enhance the training by selecting the most informative negative samples during the training process.

\begin{figure*}[!t] 
\centering 

\includegraphics[width=0.8\textwidth]{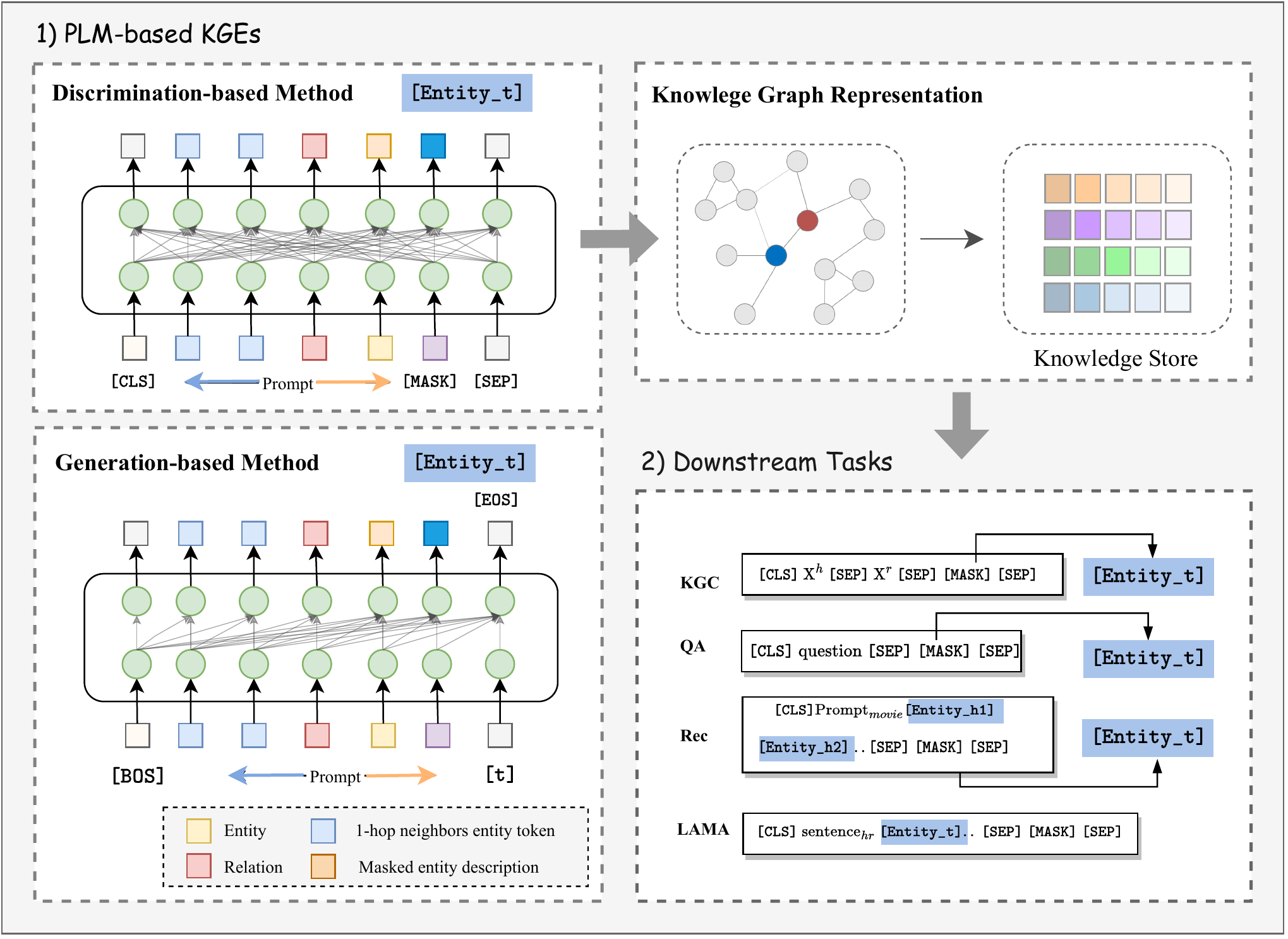} 
\caption{
PLM-based KGEs in {\ours} and those KGEs can be applied to KGC, QA, recommendation and knowledge probing.
\emph{Entity\_t} refers to the target tail entity, answer entity, recommended items, and target tail entity for different tasks, which follows the pre-train (obtain the embedding) and fine-tune paradigm (task-specific tuning).
}
\label{fig:ins}
\end{figure*}

\subsection{Unified KG Encoder}
Since {\ours} is based on PLMs, the most critical thing is to convert structural triples into plain natural language for PLMs to understand.
We introduce a unified KG encoder to represent graph structure and text semantics, supporting different types of PLM-based KGE methods.
To encode the graph structure, we sample 1-hop neighbor entities and concatenate their tokens as input for implicit structure information. 
With such a unified KG encoder, {\ours} can encode both heterogeneous graph structure and text-rich semantic information.
For the discrimination-based method, the input is built on the plain text description:
\begin{align}
\begin{split}
    X_{\text{hr pair}} &= \texttt{[CLS]}\ X^h \texttt{[SEP]} \ X^r \ \texttt{[SEP]} \\
    X_{\text{tail}} &= \texttt{[CLS]} \  X^t  \ \texttt{[SEP]}.
\end{split}
\end{align}

where $X^h$, $X^r$, and $X^t$ refer to the text sequence of the head entity, relation, and tail entity, respectively.
Referring to some prompt learning methods like $k$NN-KGE \cite{DBLP:journals/corr/abs-2201-05575}, we represent entities and relations in KG with special tokens (See \S \ref{sec:dis}) and obtain the input as:

\begin{equation}
\resizebox{.48\textwidth}{!}{$
    X = \texttt{[CLS]} X^h \texttt{[Entity h]} \ \texttt{[SEP]} \ X^r \ \texttt{[SEP]} \ \texttt{[MASK]} \ \texttt{[SEP]},
    $}
\end{equation}
where \texttt{[Entity h]} represents the special token to the head entity.

For the generation-based model, we leverage the tokens in $X^h$ and $X^r$ to optimize the model with the label $X^t$.
When predicting the head entity, we add a special token \texttt{[reverse]} in the input sequence for reverse reasoning.

\let\mc\multicolumn

\begin{table*}[!thbp]
\centering
\resizebox{.7\textwidth}{!}{
\begin{tabular}{llll}
\toprule
Model & PLM & Support Tasks & Complexity \\
\midrule 
KGBERT \cite{DBLP:journals/corr/abs-1909-03193} & MLM & KGC & $\mathcal{O}( |L|^2|\mathcal{E}|^2|\mathcal{R}|)$\\
StAR \cite{DBLP:conf/www/WangSLZW021} & MLM & KGC & $\mathcal{O}(|L/2|^2|\mathcal{E}|(1+|\mathcal{R}|) )$\\
SimKGC \cite{DBLP:conf/acl/0046ZWL22} & MLM & KGC & $\mathcal{O}(|L/2|^2|\mathcal{E}|(1+|\mathcal{R}|) )$\\
$k$NN-KGE \cite{DBLP:journals/corr/abs-2201-05575} & MLM & KGC, LAMA & $\mathcal{O}(|L|^2|\mathcal{E}||\mathcal{R}|) $\\
KGT5 \cite{DBLP:conf/acl/SaxenaKG22} & Seq2Seq & KGC, QA & $\mathcal{O}(|L/2|^3 |\mathcal{E}| |\mathcal{R}| )$\\
GenKGC \cite{DBLP:conf/www/XieZLDCXCC22} & Seq2Seq & KGC, QA & $ \mathcal{O}(|L/2|^3 |\mathcal{E}| |\mathcal{R}|)$ \\
\bottomrule
\end{tabular}
}
\caption{Comparison of different methods based on small PLMs. 
$|L|$ is the length of the triple description. $|L/2|$ can be seen as the length of entity tokens. $|\mathcal{E}|$ and $|\mathcal{R}|$ are the numbers of all unique entities and relations in the graph respectively.
}
\label{tb:time}
\end{table*}

\subsection{Model Hub}

As shown in Figure \ref{fig:ins} and Table \ref{tb:time}, \textbf{LambdaKG} consists of a \texttt{Model Hub} which supports many representative PLM-based KGE methods, mainly follow the two major paradigms of discrimination-based methods and generation-based methods as:

\paragraph{Discrimination-based methods}
\label{sec:dis}
There are three kinds of models based on the discrimination method: 
the first one  (e.g., KG-BERT \cite{DBLP:journals/corr/abs-1909-03193}, PKGC \cite{DBLP:conf/acl/LvL00LLLZ22}) utilizes a single encoder to encode triples of KG with text description;
another kind of model (e.g., StAR \cite{DBLP:conf/www/WangSLZW021}, SimKGC \cite{DBLP:conf/acl/0046ZWL22}) leverages siamese encoder (two-tower models) with PLMs to encode entities and relations respectively.
For the first kind, the score of each triple is expressed as:
\begin{equation}
    \text{Score}(h,r,t)= \text{TransformerEnc}(X^h,X^r,X^t),
\end{equation}
where \text{TransformerEnc} is the BERT model followed by a binary classifier.
However, these models have to iterate all the entities calculating scores to decide the correct one, which is computation-intensive, as shown in Table \ref{tb:time}.
In contrast, two-tower models like StAR \cite{DBLP:conf/www/WangSLZW021} and SimKGC \cite{DBLP:conf/acl/0046ZWL22} usually encode $\langle h,r\rangle$ and $t$ to obtain the embeddings.
Then, they use a score function to predict the correct tail entity from the candidates, denoted by:
\begin{equation}
    \text{Score}(\langle h,r\rangle, t) = \cos(e_{\langle h,r\rangle}, e_{t}).
\end{equation}

The final kind of model, e.g., $k$NN-KGE \cite{DBLP:journals/corr/abs-2201-05575}, utilizes masked language modeling for KGE, which shares the same architecture as normal discrimination PLMs. 
Note that there are two modules in the normal PLMs: a \textit{word embedding layer} to embed the token ids into semantic space and an \textit{encoder} to generate context-aware token embedding.
Here, we take the masked language model and \textbf{treat entities and relations as special tokens} in the ``word embedding layer''.
As shown in  Figure \ref{fig:ins}, the model predicts the correct tail entity with the sequence of the head entity and relation token and their descriptions.
For the entity/relation embedding, we freeze the \textit{encoder layer}, only tuning the \textit{entity embedding layer}, to optimize the loss function:
\begin{equation}
\resizebox{.48\textwidth}{!}{$\mathcal{L}=-\frac{1}{\left| \mathcal{E} \right|} \sum\limits_{e_j \in \mathcal{E}} \mathbb{I}(e_j=e_i) \log p\left(\texttt{[MASK]}=e_j \mid X^i ; \Theta \right)$},
\label{loss}
\end{equation}

where $\Theta$ represents the parameters of the model, $X^i$ and $e_i$ is the description and the embedding of entity $i$.

\paragraph{Generation-based methods}

Generation-based models formulate KG completion or other KG-intensive tasks as sequence-to-sequence generation.
Given a triple with the tail entity missing $(h, r, ?)$, models are fed with $\langle X^h, X^r\rangle$ and then output $X^t$.
In the training procedure, generative models maximize the conditional probability:
\begin{equation}
\resizebox{.48\textwidth}{!}{$
\text{Score}(h, r, t)= \prod_{i=1}^{|X^t|}p(x^t_i|x^t_1, x^t_2, ..., x^t_{i-1};\langle X^h, X^r \rangle).    
$}
\end{equation}

To guarantee the consistency of decoding sequential schemas and tokens in KG, GenKGC \cite{DBLP:conf/www/XieZLDCXCC22} proposes an entity-aware hierarchical decoder to constrain $X^t$.
Besides, KGT5 \cite{DBLP:conf/acl/SaxenaKG22} proposes to pre-train generation-based PLMs with text descriptions for KG representation.

\begin{figure}[!t]

\centering 
\includegraphics[width=0.48\textwidth]{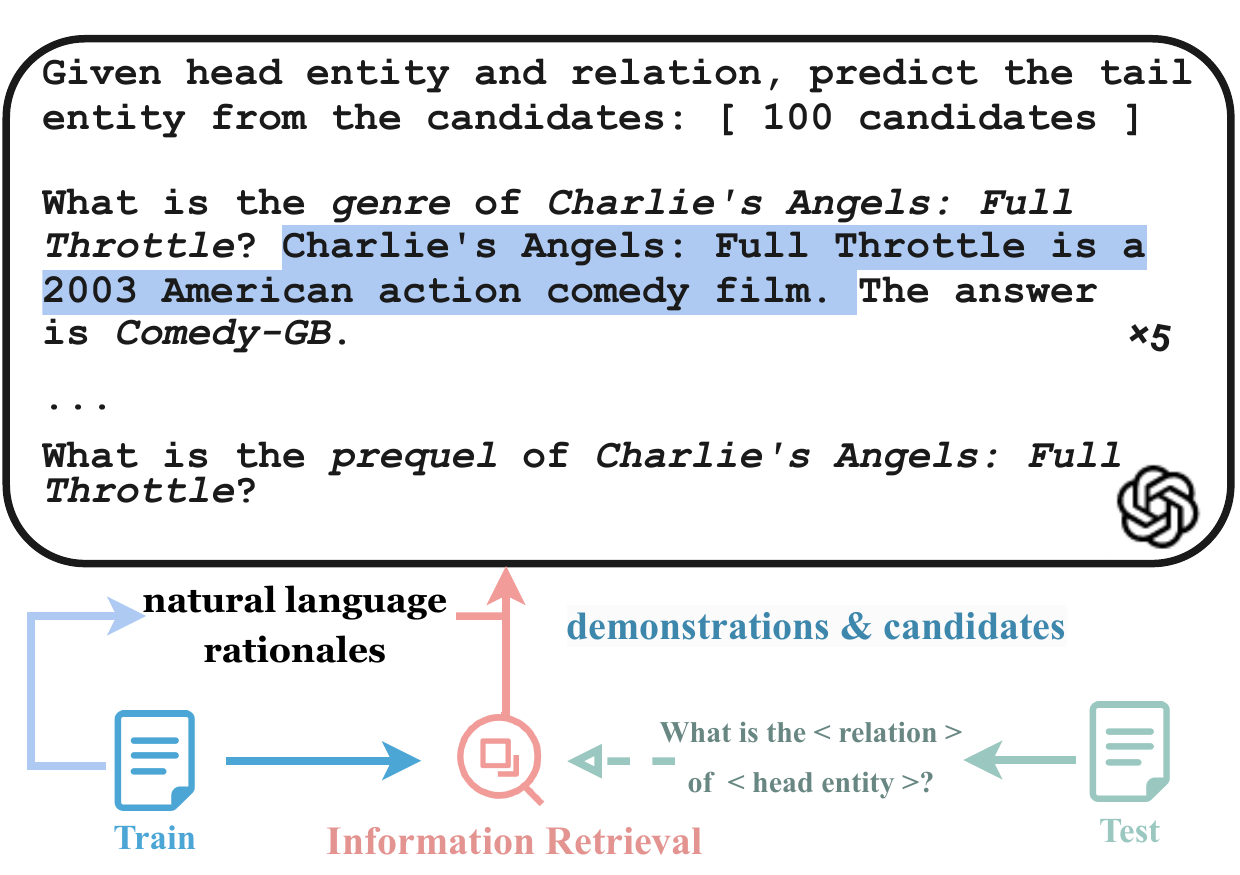} 
\caption{LLM-based KGC. The prompt comprises
three components, namely the task description with
candidates, demonstrations, and test information.
}
\label{fig:prompt}
\end{figure}

\paragraph{LLMs}
We further apply the LLMs, namely GPT-3 and ChatGPT, to assess their effectiveness in KGE (KGC with link prediction).
Generative LLMs allow the KGC task to be framed as input sentences containing header entities and relations, making it easier for the model to generate sentences with tail entities. 
A well-designed prompt can improve the performance of LLMs, and prior studies indicate incorporating in-context learning can improve accuracy and ensure consistent output. 
Thus, we adopt a similar approach that the prompt comprises three components: task description with candidates, demonstrations, and test information.

As shown in Figure \ref{fig:prompt}, we employ information retrieval (BM25) to select the top 100 most relevant entities from the training set as candidates.
Likewise, the prompt's demonstrations utilize the top-5 most similar instances, which assist the model in comprehending the task more effectively.
Furthermore, taking inspiration from the Chain-of-Thought (CoT) method in reasoning tasks, we utilize \texttt{natural language rationales} to improve the model's capacity to reason and explain predictions, ultimately improving its overall performance in KGC tasks.
Comparatively, the prompt used for ChatGPT solely utilizes a few demonstrations and test data with these strategies.

\subsection{Pluggable KGE for Downstream Tasks}

We introduce the technical details of applying KGE to downstream tasks as shown in Figure \ref{fig:ins}.
For knowledge graph completion, we feed the model with the textual information $\langle X^h, X^r\rangle$ of the head entity and the relation, then obtain the target tail entity via mask token prediction.
For question answering, we feed the model with the question written in natural language concatenated with a \texttt{[MASK]} token to obtain the special token of the target answer (entity).
For recommendation, we take the user's interaction history as sequential input \cite{DBLP:conf/cikm/SunLWPLOJ19} with entity embeddings and then leverage the mask token prediction to obtain recommended items.
For the knowledge probing task, we adopt entity embedding as additional knowledge following PELT \cite{DBLP:conf/acl/YeL0S022}.

\section{System Usage}
The proposed system can be used in three scenarios.  
First, users can utilize {\ours} to obtain \textbf{PLM-based KGE for knowledge discovery}.
\texttt{LitModel} serves as the training of link prediction task class and fit for all models in Model Hub.
Users can choose proper models in \texttt{ModelModule} and specific metrics in \texttt{DataModule} to train models to obtain the embedding in the KGs.
Moreover, users can utilize {\ours} \textbf{PLM-based KGE for downstream tasks}.
We provide various prompts to obtain the knowledge (entity) embedding in KGs for downstream tasks.
For different tasks, we design different base classes for users to efficiently implement their own tasks.
Finally, we provide an online interactive demo for PLM-based KGE at \url{https://zjunlp.github.io/project/promptkg/demo.html}.

\section{Evaluation}

 \begin{table}
  \centering
  \resizebox{.478\textwidth}{!}{
    \begin{tabular}{cclcc}
    \toprule
    \textbf{Task} & \textbf{Dataset} & \textbf{Method} & \textbf{hits1} & \textbf{MRR} \\
    \midrule
    \multicolumn{1}{c}{\multirow{12}[6]{*}{\textbf{KG Completion}}} & \multirow{6}[2]{*}{WN18RR} & KG-BERT${^\Diamond}$ & 4.1 & 21.6  \\
    & &  StAR${^\Diamond}$ & 24.3 & 40.1\\
    & &  SimKGC & 42.5 & 60.8\\
    & &  KGT5 & 17.9 & - \\
    & &  GenKGC & 39.6 & -\\
    & &  $k$NN-KGE & 52.5& 57.9\\

\cmidrule{2-5}      
& \multirow{6}[2]{*}{FB15k-237} & KG-BERT$^{\Diamond}$ &  - & -\\
    & &  StAR${^\Diamond}$ & 20.5 & 29.6 \\
    & &  SimKGC & 22.6 & 30.1\\
    & &  KGT5 & 10.8 & -\\
    & &  GenKGC & 19.2 & -\\
    & &  $k$NN-KGE & 28.0 & 37.3\\
    \midrule
    \multirow{3}[1]{*}{\textbf{Question  Answering}} & \multirow{3}[1]{*}{MetaQA} & GT query${^\Diamond}$ & 63.3 & - \\
      &   & PullNet${^\Diamond}$  & 65.1  & -  \\
      &   &    KGT5 & 67.8  & - \\
    \midrule
    \multirow{2}[1]{*}{\textbf{Recommendation}} & \multirow{2}[1]{*}{ML-20m} & BERT4Rec${^\Diamond}$ & 34.4 & 47.9  \\
    & & {\ours} & 37.3 & 50.5 \\
    \midrule
    \multirow{12}[4]{*}{\textbf{Knowledge Probing}} & \multirow{3}[2]{*}{TREx} & BERT & 28.6 & 37.7  \\
    & & RoBERTa & 19.9 & 27.8 \\
    & & {\ours} (RoBERTa) & 22.1 & 29.8 \\
    \cmidrule{2-5}
    & \multirow{3}[2]{*}{Squad} & BERT & 13.2 & 23.5  \\
    & & RoBERTa & 13.4 & 24.6 \\
    & & {\ours} (RoBERTa) & - & - \\
    \cmidrule{2-5}
    & \multirow{3}[2]{*}{Google RE} & BERT & 10.3 & 17.3 \\
    & & RoBERTa & 7.6 & 12.8 \\
    & & {\ours} (RoBERTa) & 8.1 & 14.2 \\
    \bottomrule
    \end{tabular}
    }
         \caption{
  Hits1 and MRR (\%) results on KGC, question answering, recommendation and  knowledge probing tasks. 
  $\Diamond$ refers to the results from origin papers.}
    
    \label{tab:sdm}
\end{table}
 \begin{figure}[!t]

\centering 
\includegraphics[width=0.5\textwidth]{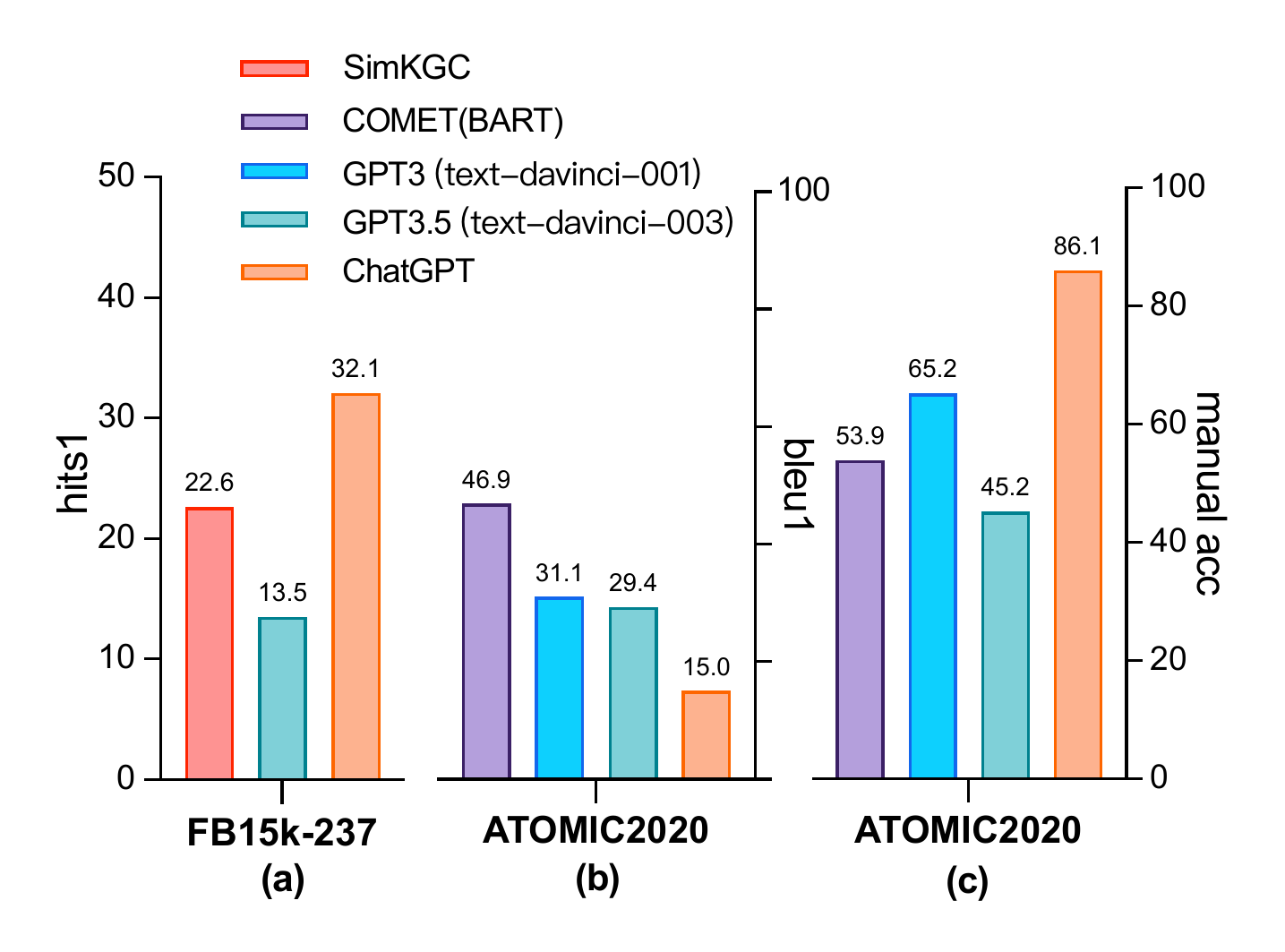} 

\caption{Results on small and large LMs. (a) hit@1 scores on FB15k-237. (b) BLEU-1 scores on ATOMIC2020. (c) Accuracy scores on ATOMIC2020 by manual evaluation.
}
\label{fig:gpt}
\end{figure}


\subsection{Knowledge Graph Completion}

For the KG completion task with small PLMs, we conduct link prediction experiments on two datasets WN18RR \cite{DBLP:conf/aaai/DettmersMS018}, and FB15k-237 \cite{DBLP:conf/emnlp/ToutanovaCPPCG15}.
From Table \ref{tab:sdm}, we observe that the discrimination-based method SimKGC \cite{DBLP:conf/acl/0046ZWL22} (previous state-of-the-art) achieves higher performance than other baselines.
Generation-based models like KGT5 \cite{DBLP:conf/acl/SaxenaKG22} and GenKGC \cite{DBLP:conf/www/XieZLDCXCC22} also yield comparable results and show potential abilities in KG representation.

\paragraph{Small vs. Large LMs}
We adopt GPT-3/3.5 (\texttt{text-davinci-001/003} and ChatGPT) for evaluation and assessment through the interfaces provided by OpenAI.
The evaluation of ChatGPT is conducted on 224 instances, with each relation in the test set.
As shown in Figure \ref{fig:gpt}(a), ChatGPT demonstrates better performance, while \texttt{text-davinci-003} exhibits a slight gap.
The experiment has reaffirmed the capability of LLMs in capturing semantic similarities and regularities among entities, thereby allowing for precise predictions of missing links in knowledge graphs.

\begin{figure}[!t]

\centering 
\includegraphics[width=0.5\textwidth]{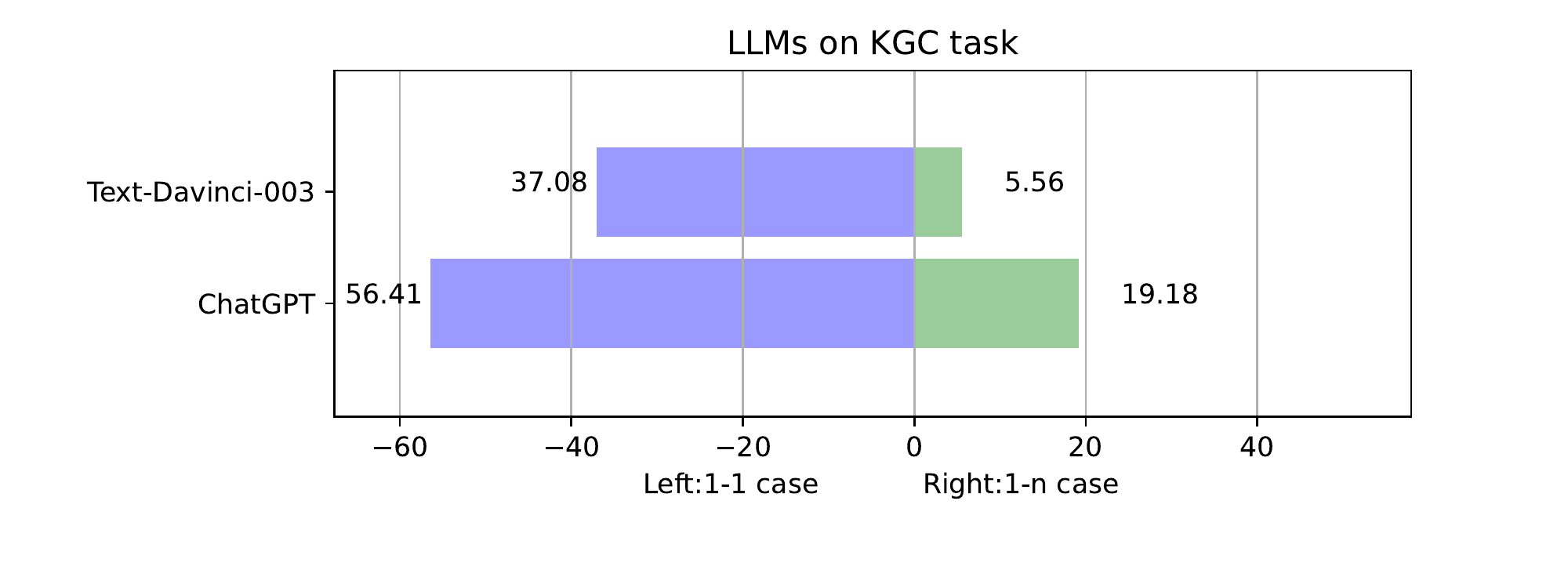} 

\caption{hit@1 of ChatGPT and \texttt{text-davinci-003} in FB15k-237. 
}
\label{fig:fig_1}
\end{figure}

In cases where one head entity and relation pair correspond to one or multiple tail entities (1-1 and 1-n cases), we conducted a detailed analysis. Notably, the model performs significantly better in the 1-1 case compared to the 1-n case, as illustrated in Figure \ref{fig:fig_1}. 
Two potential reasons explain this disparity:
(1) In the 1-1 case, the model demonstrates a lower propensity for language understanding deviations. 
Additionally, ChatGPT's training utilizes a larger corpus, enhancing the model to generate accurate responses through analysis and reasoning.
(2) The presence of multiple correspondences poses a challenge for the model's capacity to generate informative and contextually relevant responses. 
Moreover, current evaluation metrics fail to fully capture the intricacy of the responses necessary to properly handle such questions.

We further conduct experiments on \textbf{commonsense KG completion} with ATOMIC2020 \cite{DBLP:conf/aaai/HwangBBDSBC21}.
As suggested in the paper, we sample 5,000 test queries to evaluate the models (excluding ChatGPT). 
COMET (BART) is fine-tuned through supervised learning and utilizes greedy decoding to generate answers. 
For GPT3 and ChatGPT, we provide each relation with 5 examples of heads and tails to construct prompts and evaluate them in a zero-shot setting. 
The results, as shown in Figure \ref{fig:gpt}(b), demonstrates the BLEU-1 scores on the sampled 5,000 queries, while we sample 115 (5 for each relation) queries from the test for ChatGPT. 
The results indicate that GPT-3 exhibits limited performance in the system evaluation.
After analyzing several cases, we sample 115 (5 for each relation) queries as a benchmark and apply manual scoring to evaluate models.
Figure \ref{fig:gpt}(c) depicts the accuracy scores of each model.
Our study reveals that ChatGPT is capable of generating reasonable outputs, but they are quite different from the ground truth, which accounts for the final results.

\subsection{Question Answering}
KG is known to be helpful for the task of question answering.
We apply \textbf{LambdaKG} to question answering and conduct experiments on the MetaQA dataset.
Due to computational resource limits, we only evaluate the 1-hop inference performance.
From Table \ref{tab:sdm}, KGT5 in \textbf{LambdaKG} yields the best performance. 

\subsection{Recommendation}
For the recommendation task, we conduct experiments on a well-established version ML-20m\footnote{\url{https://grouplens.org/datasets/movielens/20m/}}. 
Linkage of ML-20m and Freebase offered by KB4Rec \cite{DBLP:journals/dint/ZhaoHYDHOW19} is utilized to obtain textual descriptions of movies in ML-20m. 
With movie embeddings pre-trained on these descriptions, we conduct experiments on sequential recommendation tasks following the settings of BERT4Rec \cite{DBLP:conf/cikm/SunLWPLOJ19}. 
We notice that \textbf{LambdaKG} is confirmed to be effective for the recommendation compared with BERT4Rec.

\subsection{Knowledge Probing}
Knowledge probing \cite{DBLP:conf/emnlp/PetroniRRLBWM19} examines the ability of LMs (BERT, RoBERTa, etc.) to recall facts from their parameters.
We conduct experiments on LAMA using pre-trained BERT (\textit{bert-base-uncased}) and RoBERTa (\textit{roberta-base}) models.
To prove that entity embedding enhanced by KGs helps LMs grab more factual knowledge from PLMs, we train a pluggable entity embedding module following PELT \cite{DBLP:conf/acl/YeL0S022}.
As shown in Table \ref{tab:sdm}, the performance boosts while we use the entity embedding module.

\section{Conclusion and Future Work}
We propose \textbf{LambdaKG}, a library that establishes a unified toolkit with well-defined modules and easy-to-use interfaces to support research on using PLMs on KGs.
In the future, we will continue to integrate more models and tasks (e.g., dialogue) into the proposed library to facilitate the research progress of the KG.


\section*{Acknowledgment}

We would like to express our heartfelt gratitude to the anonymous reviewers for their thoughtful and kind comments.
This work was supported by the National Natural Science Foundation of China (No.62206246), Zhejiang Provincial Natural Science Foundation of China (No. LGG22F030011), Ningbo Natural Science Foundation (2021J190), Yongjiang Talent Introduction Programme (2021A-156-G).

\bibliography{anthology,custom}
\bibliographystyle{acl_natbib}




\end{document}